\documentclass[sigconf]{acmart}
\setcopyright{none}

\usepackage{algorithm}
\usepackage{soul}
\usepackage{courier}
\usepackage{graphicx} % Required for inserting images
\usepackage{multirow}
\usepackage{multicol}
\usepackage{rotating}
\usepackage{booktabs}
\usepackage{hyperref}
\usepackage{algpseudocode}

\AtBeginDocument{%
  \providecommand\BibTeX{{%
    Bib\TeX}}}

\copyrightyear{2023}
\acmYear{2023}
\setcopyright{acmlicensed}\acmConference[NLPIR 2023]{2023 7th International Conference on Natural Language Processing and Information Retrieval}{Dec. 15--17, 2023}{Seoul, Republic of Korea}
\acmBooktitle{2023 7th International Conference on Natural Language Processing and Information Retrieval (NLPIR 2023), Dec. 15--17, 2023, Seoul, Republic of Korea}
\acmDOI{10.1145/3639233.3639253}
\acmISBN{979-8-4007-0922-7/23/12}
\begin{document}

%%
%% The "title" command has an optional parameter,
%% allowing the author to define a "short title" to be used in page headers.
\title{Action-Item-Driven Summarization of Long Meeting Transcripts}

%%
%% The "author" command and its associated commands are used to define
%% the authors and their affiliations.
%% Of note is the shared affiliation of the first two authors, and the
%% "authornote" and "authornotemark" commands
%% used to denote shared contribution to the research.
% \author{Ben Trovato}
% \authornote{Both authors contributed equally to this research.}
% \email{trovato@corporation.com}
% \orcid{1234-5678-9012}
% \author{G.K.M. Tobin}
% \authornotemark[1]
% \email{webmaster@marysville-ohio.com}
% \affiliation{%
%   \institution{Institute for Clarity in Documentation}
%   \streetaddress{P.O. Box 1212}
%   \city{Dublin}
%   \state{Ohio}
%   \country{USA}
%   \postcode{43017-6221}
% }

\author{Logan Golia}

\affiliation{%
  \institution{Rice University}
  \country{USA}
  }
  \email{lsg3@rice.edu}

\author{Jugal Kalita}
\affiliation{%
  \institution{University of Colorado, Colorado Springs}
  \country{USA}}
\email{jkalita@uccs.edu}
% \author{Valerie B\'eranger}
% \affiliation{%
%   \institution{Inria Paris-Rocquencourt}
%   \city{Rocquencourt}
%   \country{France}
% }

% \author{Aparna Patel}
% \affiliation{%
%  \institution{Rajiv Gandhi University}
%  \streetaddress{Rono-Hills}
%  \city{Doimukh}
%  \state{Arunachal Pradesh}
%  \country{India}}

% \author{Huifen Chan}
% \affiliation{%
%   \institution{Tsinghua University}
%   \streetaddress{30 Shuangqing Rd}
%   \city{Haidian Qu}
%   \state{Beijing Shi}
%   \country{China}}

% \author{Charles Palmer}
% \affiliation{%
%   \institution{Palmer Research Laboratories}
%   \streetaddress{8600 Datapoint Drive}
%   \city{San Antonio}
%   \state{Texas}
%   \country{USA}
%   \postcode{78229}}
% \email{cpalmer@prl.com}

% \author{John Smith}
% \affiliation{%
%   \institution{The Th{\o}rv{\"a}ld Group}
%   \streetaddress{1 Th{\o}rv{\"a}ld Circle}
%   \city{Hekla}
%   \country{Iceland}}
% \email{jsmith@affiliation.org}

% \author{Julius P. Kumquat}
% \affiliation{%
%   \institution{The Kumquat Consortium}
%   \city{New York}
%   \country{USA}}
% \email{jpkumquat@consortium.net}

%%
%% By default, the full list of authors will be used in the page
%% headers. Often, this list is too long, and will overlap
%% other information printed in the page headers. This command allows
%% the author to define a more concise list
%% of authors' names for this purpose.
\renewcommand{\shortauthors}{}

%%
%% The abstract is a short summary of the work to be presented in the
%% article.
\begin{abstract}
  The increased prevalence of online meetings has significantly enhanced the practicality of a model that can automatically generate the summary of a given meeting. This paper introduces a novel and effective approach to automate the generation of meeting summaries. Current approaches to this problem generate general and basic summaries, considering the meeting simply as a long dialogue. However, our novel algorithms can generate abstractive meeting summaries that are driven by the action items contained in the meeting transcript. This is done by recursively generating summaries and employing our action-item extraction algorithm for each section of the meeting in parallel. All of these sectional summaries are then combined and summarized together to create a coherent and action-item-driven summary. In addition, this paper introduces three novel methods for dividing up long transcripts into topic-based sections to improve the time efficiency of our algorithm, as well as to resolve the issue of large language models (LLMs) forgetting long-term dependencies. Our pipeline achieved a BERTScore of 64.98 across the AMI corpus, which is an approximately 4.98\% increase from the current state-of-the-art result produced by a fine-tuned BART (Bidirectional and Auto-Regressive Transformers) model.\footnote{\href{https://github.com/logangolia/meeting-summarization}{https://github.com/logangolia/meeting-summarization}}

\end{abstract}

%%
%% The code below is generated by the tool at http://dl.acm.org/ccs.cfm.
%% Please copy and paste the code instead of the example below.
%%


%%
%% Keywords. The author(s) should pick words that accurately describe
%% the work being presented. Separate the keywords with commas.

% \received{20 February 2007}
% \received[revised]{12 March 2009}
% \received[accepted]{5 June 2009}

%%
%% This command processes the author and affiliation and title
%% information and builds the first part of the formatted document.
\maketitle

\section{Introduction}

As a result of the COVID-19 pandemic, many professional meetings and conversations have been conducted online; this also means that the transcripts of these meeting have become readily available. Minutes are accepted official records of what transpired in a meeting, and so designated personnel usually conduct the tedious process of generating meeting minutes. However, with the help of large language models (LLMs), we can automate this process and still generate factual and informative summaries.

There are two main approaches to text summarization: extractive and abstractive. Extractive summarization techniques locate the most important phrases and sentences from the input transcript and concatenate them to form a concise summary. However, the summaries generated by these techniques are usually awkward to read because of the forceful concatenation of unrelated sentences \cite{koh_empirical_2023}. Abstractive summarization techniques focus more on understanding the overall meaning of a transcript and then generating a concise summary based on the entire text. Unlike extractive summarization, abstractive summarization generates new words and phrases that were not found in the input transcript, rather than simply extracting the important phrases \cite{rennard_abstractive_2023}. Abstractive summarization is more challenging, but as expected, it leads to better summaries \cite{gupta_abstractive_2019}. As a result, meeting summarization has begun to head in this direction, and this study utilizes abstractive summarization techniques as well.

Current approaches to automating the creation of meeting minutes treat summarizing a meeting the same way they would summarize a dialogue \cite{fm_automation_2022}. However, we argue that meeting summarization is fundamentally different from dialogue summarization. Unlike a dialogue, useful meeting minutes have additional features that are often not included in the automated summary of the meeting: action items, main topics, tension levels, decisions made, etc. In this study, we focus on incorporating action items into the machine-generated summaries.

LLMs today still struggle to capture long-term dependencies in texts, and as a result, they are not very good at generating summaries for long transcripts \cite{dong_survey_2023}. The time and space complexities of these transformer-based models increase quadratically with respect to the input size \cite{vaswani_attention_2017}, and new LLMs still have strict input token limits \cite{yang_exploring_2023}. Most solutions to these problems employ linear segmentation, where the long texts are broken up into equal subsections based on token numbers, but the problem with this approach is that we inevitably interrupt ideas in the text. We build upon previous work in topic segmentation to divide the text into topical chunks before summarizing. %We also introduce a new technique for extracting action items from a text. In this study we employ a recursive approach to generating abstractive summaries of long meeting transcripts that are driven by the action items presented in the meeting.  

In summary, current solutions to the problem of automatically generating meeting minutes given the transcript of the meeting produce general and vague summaries. In addition, there is a lack of effective topic segmentation methods for meeting summarization. This study outlines a novel method of utilizing topic segmentation and recursive summarization to generate action-item-driven abstractive summaries of long meeting transcripts.

Our main contributions are threefold:

1) We develop three novel topic segmentation algorithms, in which the best outperforms the summarization performance provided by linear segmentation by 1.36\% in terms of the BERTScore metric;

2) We develop our own effective action-item extraction algorithm;

3) Our novel parallel and recursive meeting summarization algorithm properly generates action-item-driven summaries and improves upon the performance of current state-of-the-art models by approximately 4.98\% in terms of the BERTScore metric.

\section{Related Work}
In this section, we discuss previous methods employed in meeting summarization and provide motivation for our novel techniques.

\subsection{Recursive Summarization}
One way in which meeting summarization differs from dialogue summarization is that meeting transcripts are generally long, and as explained earlier, transformer-based models struggle with larger input sizes. As a result, it has been proven effective to divide long documents into multiple parts, summarize each component, and then combine the summaries back together in a recursive approach. The recursive algorithm described in this paper is inspired by the method proposed by \citet{wu_recursively_2021}, which was used to summarize long books. The methods proposed by \citet{shinde_automatic_2022} and \citet{yamaguchi_team_2021} are not truly recursive because after they combine the sectional summaries back together, the final summary is never fed back into the summarization model. Instead, they perform argument mining on the resulting chunk of the combined summaries. We propose a truly recursive approach and achieve state-of-the-art results with this technique. 

\subsection{BART Model for Meeting Summarization}
While there do exist more powerful dialogue summarization models such as DialogLM \cite{zhong_dialoglm_2022} and Summ$^N$ \cite{zhang_summn_2022}, we use the BART (Bidirectional and Auto-Regressive Transformers) model \cite{lewis_bart_2020} due to its speed and high performance in long document summarization tasks \cite{koh_empirical_2023}. In addition, there has been previous research in assessing different topic segmentation methods on the BART model, so this allows us to evaluate our techniques.  

\subsection{AMI Dataset}
The AMI dataset is a large meeting corpus consisting of 137 scenario-driven meetings and their corresponding summaries \cite{AMI}. Even though the scenarios are artificial, the way in which the actors choose how to interact with each other is spontaneous. The realistic meeting conversations combined with the fact that there are 137 different long meeting transcripts makes the AMI corpus an ideal dataset on which to test our techniques on. 

\subsection{Current Segmentation Techniques}
There are several techniques to divide meeting transcripts into multiple parts, but none have actually been able to improve summarization results when compared to the simplest technique, linear segmentation. Linear segmentation is the process of dividing the meeting transcript into parts solely based on token number, including a preset number of tokens in each chunk. The state-of-the-art results on summarizing the AMI corpus using the BART model are achieved through this technique by \citet{shinde_automatic_2022}. \citet{shinde_automatic_2022} attempted to use two additional topic segmentation techniques, Depth-Scoring \cite{solbiati_unsupervised_2021} and TextTiling \cite{hearst_texttiling_1997}, but neither were able to improve upon the results obtained by linear segmentation. \citet{yamaguchi_team_2021} also introduces a novel technique for topic segmentation using a Longformer+LSTM model to predict whether a sentence is the start of a new topic, in the middle of a topic, or outside of a particular topic. However, their summarization results were significantly lower than those achieved by \citet{shinde_automatic_2022}. We propose three novel segmentation techniques that outperform linear segmentation. 

\subsection{Evaluation Metrics}
ROUGE's F1 scores are the most popular metrics in evaluating machine-generated summaries \cite{lin_rouge_2004}. However, ROUGE scores have many flaws since they focus solely on the lexical overlap between the machine-generated summaries and the human reference summaries rather than their semantic similarity \cite{fabbri_summeval_2021}. As a result, BERTScore, which measures the semantic similarity between the machine-generated summaries and the reference summaries has been growing in popularity \cite{rennard_abstractive_2023}. We employ the BERTScore metric as well, since it has been shown to achieve higher correlations with human judgment on the quality of a machine-generated summary compared to ROUGE \cite{zhang_bertscore_2020}.

% \section{3 Problem Statement}
% Current solutions to the problem of automatically generating meeting minutes given the transcript of the meeting produce very general and vague summaries. In addition, there is a lack of effective topic segmentation methods in the field of meeting summarization. This study outlines a novel method of utilizing topic segmentation and recursive summarization to generate action item driven abstractive summaries of long meeting transcripts.

\section{Approach}
In this section, we dive deeper into our recursive algorithm for generating action-item-driven meeting summaries. We also explore the lower-level techniques that were necessary to improve state-of-the-art results and provide motivation for these design decisions along the way.
\subsection{Divide-and-conquer}
As described in our "Introduction" and "Related Works" sections, the first step to summarizing long meeting transcripts is to break them up, so we can summarize each chunk. We propose three simple but effective topic segmentation techniques that were able to generate more truthful and concise summaries when compared to linear segmentation.
\subsubsection{Chunked Linear Segmentation}
When we ran our model using linear segmentation (splitting the text based on a preset token number across all chunks), we noticed that points were often misunderstood and repeated because we were creating separate chunks in the middle of a speaker's formulation of one idea; let us call each speaker's contiguous dialogue a "turn." Thus, we first employed a simple technique inspired by linear segmentation where we maximize the number of tokens in each chunk, adding turns until we reach a preset token number, whilst ensuring that no speaker's turn is interrupted.
\subsubsection{Simple Cosine Segmentation}
The second technique we created is based upon chunked linear segmentation, but also upon the cosine similarity of the MPNet embeddings, a state-of-the-art sentence embedding model \cite{song_mpnet_2020}, for each turn. For each turn, we compute its MPNet embedding and calculate its cosine similarity with the MPNet embedding of the previous turn. If the cosine similarity of the embeddings is greater than 0, we simply add this turn to the current chunk. If the cosine similarity of the embeddings is less than or equal to 0, we define the current turn to be the beginning of a new topic and start a new chunk. 

We choose a similarity threshold of 0 to signify the start of a new topic after experimenting with different values and manually inspecting the quality of the resulting summaries, as well as evaluating the resulting summaries with the ROUGE and BERTScore metrics. This value of 0 also makes sense in theory because it means that the two consecutive turns are more semantically dissimilar than they are similar. This leads to better results because we do not want to split the transcript into too many topics, and instead favor large topics; we generally want to keep as much text intact as possible, so the summarization model has enough context to generate a quality summary. 
% We do not want to generate too many independent summaries for each topic that have little relation to each other and then combine these little summaries together. In our testing, this proved to be a very ineffective approach because each sectional summary had little context of the surrounding text to work with, and as a result, the resulting overall summary was very confusing to read. 
This is also why topic segmentation for summarization is very different from typical topic segmentation because we do not want to create chunks at every little topic change. In fact, when we increased our similarity threshold from 0 to just 0.2, our BERTScores and ROUGE-L scores both decreased by $>1\%$ which is very significant for summarization tasks.

It is also important to note that when splitting based on some cosine similarity threshold, there is a risk that no new chunks will be created for over 1024 consecutive tokens, which is the max input token limit for the BART model \cite{obonyo_exploring_2022}. Therefore, as we move through the turns and add them to the existing chunk, we check to ensure that adding the current turn will not make the current chunk greater than 1024 tokens. If this does happen, we create a new chunk/topic beginning with this turn, regardless of this turn's cosine similarity with the previous turn.
\subsubsection{Complex Cosine Segmentation}
We noticed a recurring problem when inspecting the topic chunks that were being created by the previous method. Sometimes a person would utter something meaningless, and that would compose their entire turn (e.g. "Bob: Ummm"). As a result, this turn would often have a very low cosine similarity with the previous turn, and a new topic/chunk would be created. The simplest solution to this problem would be to remove all redundant and meaningless utterances in the pre-processing stage. The problem with this approach is that even if we somehow managed to hard code the regular expressions in order to remove all of the "meaningless" turns, there are still lots of cases where a speaker will say something completely unrelated to the current topic (e.g. "Bob: Let us go grab ice cream after this"), but then they will resume talking about the original topic. In this case, we would not want to create a new topic. In order to achieve this, we take the same approach used in "simple cosine segmentation", except we recalculate the MPNet embedding of the entire current chunk before comparing its cosine similarity to the MPNet embedding of the following turn. This mitigates the effect of "meaningless" turns, particularly consecutive "meaningless" turns, since they will have less impact on the MPNet embedding of the chunk we are comparing the next turn to. Please refer to Algorithm 1 for further details.

\begin{algorithm*}
\caption{Complex Cosine Segmentation(string text, int similarityThreshold, int maxTokens)}
\begin{algorithmic}[1]
\State $turns \gets$ $text$ split by speaker
\State $model \gets$ sentence embedding model
\State $tokenizer \gets$ tokenizer used by summarization model
\State $processedChunks \gets$ list with the first sentence from $turns$
\For{$i$ in range(1, $len(turns$))} \Comment{Iterate through the turns}
    \State $curChunkEmbedding \gets model.encode(processedChunks[-1])$
    \State $nextSpeakerEmbedding \gets model.encode(turns[i])$
    \State $similarity \gets$ $cosineSimilarity(curChunkEmbedding, nextSpeakerEmbedding)$ \Comment{Compute similarity}
    \State $newChunk \gets processedChunks[-1]$ + $turns[i]$
    \State $newNumTokens \gets tokenLen(tokenizer(newChunk))$
    \If{$similarity > similarityThreshold$ and $newNumTokens \leq maxTokens$}
        \State $processedChunks[-1] \gets newChunk$ \Comment{Add turn to the current chunk}
    \Else 
        \State append $turns[i]$ to $processedChunks$ \Comment{Start a new chunk}
    \EndIf
\EndFor
\State \Return $processedChunks$ \Comment{A list of topic-based chunks of $text$}
\Description[Complex Cosine Segmentation algorithm]{Our novel "complex cosine segmentation" algorithm as described in Section 3.1.3}
\end{algorithmic}
\end{algorithm*}

\subsection{Generating the General Sectional Suammries}
Once we have divided the original text into chunks, the next step is to generate a general abstractive summary for each chunk. Our approach to solve this problem involves fine-tuning Meta’s BART model \cite{lewis_bart_2020}, a pre-trained large language model, on dialogue datasets to generate general summaries of a meeting. We elect to use a BART model since its bidirectional encoder and auto-regressive decoder have been shown to better understand the full semantics of a text and generate coherent summaries \cite{lewis_bart_2020}. Specifically, we used a BART model fine-tuned on the XSUM \cite{narayan_dont_2018} and SAMSUM \cite{gliwa_samsum_2019} datasets to generate the general summaries for each chunk. These are widely used dialogue datasets for training dialogue summarization models \cite{feng_survey_2022}. They are also the same datasets on which \citet{shinde_automatic_2022} fine-tuned their model, so we can better compare our results.

% In addition, we noticed that since each general sectional summary is independent of one another, they can be generated in parallel. To the best of our knowledge, we are the first to incorporate parallelism in the divide-and-conquer summarization algorithm as see in Algorithm 3. Let totalT = num tokens in the transcript and maxT = input token limit of summarization model. The worst-case (lower bound) theoretical parallel speedup would be ceiling(totalT / maxT)x. The worst-case speedup comes from the scenario where the similarity threshold in our topic segmentation algorithms are never triggered and our chunks formed then approach those that would have been created by linear segmentation. For example, if our meeting transcript contained 4,757 tokens, which is the average number of tokens in an AMI transcript (ZHU ET AL), and our summarization model had an input token limit of 1024, which is the input token limit of the BART model (CITATION), the worst-case speedup would be 5x which is still a great improvement.  
In addition, we noticed that since each general sectional summary is independent of one another, they can be generated in parallel. To the best of our knowledge, we are the first to incorporate parallelism in the divide-and-conquer summarization algorithm as seen in Algorithm 3. 

% Let $totalT$ denote the total number of tokens in the transcript and $maxT$ represent the input token limit of the summarization model. The worst-case (lower bound) theoretical parallel speedup, $S$, would be calculated as 
% \begin{equation}
% S = \left\lfloor\frac{{totalT}}{{maxT}}\right\rfloor
% \end{equation}
% The worst-case speedup scenario arises when the similarity threshold in our topic segmentation algorithm is never triggered. In this case, the chunks that we form become similar to those that would have been created by a linear segmentation method. 

% For example, if our meeting transcript contained 4,757 tokens, which is the average number of tokens in an AMI transcript \cite{zhu_hierarchical_2020}, and our summarization model had an input token limit of 1024, which is the input token limit of the BART model \cite{obonyo_exploring_2022}, the worst-case speedup would be calculated as 
% \begin{equation}
% S = \left\lfloor \frac{{4757}}{{1024}} \right\rfloor = 4
% \end{equation}
% which is still a great improvement.

\subsection{Action-Item Extraction}
Another very important component of any good meeting summary is what each participant has accomplished and what they need to accomplish before the next meeting; so for each chunk of text, we need to extract the action items. Although recording action items is an important part of many meeting summaries, the issue has been ignored in prior work. This problem was first introduced by \citet{cohen_automatic_2021}, but little progress has been made since. To solve this, we use a public dataset\footnote{\url{https://github.com/kiransarv/actionitemdetection/blob/master/dataset}} from a GitHub repository that contains 2750 dialogue statements as well as corresponding labels for whether a statement contains action items or not. We then fine-tune a BertForSequenceClassification\footnote{\url{https://huggingface.co/docs/transformers/v4.31.0/en/model_doc/bert\#transformers.BertForSequenceClassification}} model (a BERT model \cite{bert} with a linear layer on top for classification) on this dataset to classify the action items in the original meeting transcript. This training method proved effective with a classification accuracy of 95.4\% on the test dataset. However, this process alone is not enough to extract the key action items from a text. This method alone identifies which sentences contain action items, but it does not extract the underlying ideas. For example, a sentence identified as an action item can be "you need to do that before the next meeting." This is indeed an action item, but it doesn't actually contain any useful information; there are too many pronouns and not enough context. In the next sections, we discuss existing methods to solve this problem, explain their limitations, and present our own technique.  

\subsubsection{Coreference Resolution}
We first employed widely used state-of-the-art methods and models for coreference resolution to convert the sentences that were classified as action items into more context-rich statements. We employed libraries such as Stanford CoreNLP \cite{clark_improving_2016} and NeuralCoref\footnote{\url{https://github.com/huggingface/neuralcoref}} (an extension of the spaCy library), but were not satisfied by the results. Not only were the pronouns not always resolved for larger text inputs, but we realized that coreference resolution alone was not enough. Even if the pronouns were resolved, this was often not enough context to completely understand the sentence containing the action item. For example, the sentence "you need to do that before the next meeting" may be converted to "Jake needs to fix the website before the next meeting" after coreference resolution. This is better, but it is still not enough information for Jake to read this sentence in the meeting minutes and understand what needs to be done.
\subsubsection{Context Resolution}
In this paper, we develop a technique to solve this lack-of-context problem which we call "neighborhood summarization." Once we find a sentence that has been identified as an action item, we find its "neighborhood." We define a sentence's neighborhood as the three sentences before the sentence, the sentence itself, and the two sentences after the sentence. We use all six of these sentences as inputs into the same BART summarization model that we used to generate the sectional summaries, and we are left with a rephrased version of the sentence containing the action item. We believe the reason this technique works so well is because the reference summaries in the dialogue datasets that our BART model is fine-tuned on are naturally action-item driven, to some extent. To use the same example, this neighborhood summarization technique can convert a sentence that has been identified as an action item, "you need to do that before the next meeting", into a context-rich rephrasing, "Jake needs to fix the menu button on the website because our users are complaining that it does not work half the time." 

We choose three sentences before and two sentences after for our neighborhood after experimenting with different values and inspecting the quality of the resulting summaries ourselves. Any smaller of a neighborhood, and we found that there was not enough context in the resulting summary. Any larger of a neighborhood, and the summary often did not revolve around the action item and instead addressed other parts of the input text that were not relevant for this particular action-item extraction task. It makes sense that we would need more sentences before the action item than after it since most pronoun references and necessary context would be provided before a sentence that depends on it. However, since this is a dialogue summarization task, and there are many anomalies when people speak, sentences after the action item are still necessary to include in the neighborhood in the event that additional pronoun references or context comes after. Note that there are edge cases, for example when an action item is located at the very beginning or end of a chunk, so please see Algorithm 2 for more details. 

\begin{algorithm*}
\caption{Action-Item Extraction(string text)}
\begin{algorithmic}[1]
\State $model \gets$ action item classifier
\State $tokenizer \gets$ BERT tokenizer
\State $actions \gets$ empty string
\State $sentences \gets$ $text$ split by sentence
\For{$index$, $sentence$ in enumerate($sentences$)} \Comment{Iterate through the sentences}
    \State $inputs \gets tokenizer(sentence)$
    \State $predictedClass \gets model(inputs)$
    \If{$predictedClass = 1$} \Comment{Class 1 indicates $sentence$ is an action item}
        \State $neighborhood \gets$ empty string
        \State $startIndex \gets max(0, index - 3)$
        \State $endIndex \gets min(len(sentences), index + 3)$
        \For{$neighborIdx$ in range($startIndex$, $endIndex$)}
            \State $neighborhood$ += $sentences[neighborIdx]$
        \EndFor
        \State $actions$ += $generalSum(neighborhood)$ \Comment{Summarize the neighborhood}
    \EndIf
\EndFor
\State \Return $actions$ \Comment{A string containing the context-rich action items found in $text$}
\Description[Action-Item Extraction algorithm]{Our novel "action-item extraction" algorithm as described in Section 3.3.2}
\end{algorithmic}
\end{algorithm*}

We append the action items with context from a given chunk to the end of the general summary for this same chunk. This way, we keep the summaries and action items that are derived from the same pieces of text together. Then we pass this entire text (summary + action items) into the same BART summarizer. We found that this technique helps condense the summary as well as improve the coherence of the resulting summary for each chunk.

\subsection{Combining Summaries and the Recursive Case}
Now that we have generated summaries for each chunk, containing information regarding both the general summary and the action items, we will generate an abstractive summary again based on all of the sectional summaries combined together in a recursive approach. If we append the sectional summaries together, and the number of tokens in this entire chunk of text is less than 1024, we pass this entire chunk of summaries into the same BART summarizer again; in essence, we are summarizing the summaries. However, if this entire chunk of summaries contains more than 1024 tokens, then we fall into the recursive case where we pass this entire chunk of summaries back into the entire function as if it is a meeting transcript. We explored other techniques to fluidly combine the summaries together, but we found that using the BART summarizer achieved the best results. For example, we attempted to use an existing RoBERTa model \cite{liu_roberta_2019} that was fine-tuned on a sentence fusion dataset known as DiscoFuse \cite{rothe_leveraging_2020}. However, this technique did not prove effective because the resulting summaries were often very long and contained repetitions. We tried solving this problem by tuning the BART summarizer model to generate shorter sectional summaries, so the resulting chunk of all the summaries appended together would be shorter, but the sentence fusion models still did not prove effective in generating grammatically correct and coherent final summaries. This is a very challenging task if approached from a sentence fusion perspective, howerver, we approached this problem as simply another summarization task; the fine-tuned BART summarizer proved very effective at this task by removing repetitions between the sectional summaries and generating very informative, coherent, and concise summaries as seen in our results table. 

\begin{algorithm*}
\caption{Action-Item-Driven Summary(string text, bool first, int maxTokens)}
\begin{algorithmic}[1]
\State $tokenizer \gets$ tokenizer used by summarization model
%\State $text \gets \text{preProcessText}(text)$

\If{$first = True$}
    \State $chunks \gets \text{topicalChunksBySpeaker}(text)$ \Comment{Split text into topic-based chunks}
\Else
    \State $chunks \gets \text{topicalChunksBySentence}(text)$
\EndIf
\State $chunkSums \gets \text{array\  with\ size\ of\ len(chunks)}$ 

\ForAll{$index \in \text{range}(0, \text{len}(chunks))$} \Comment{Summarize each chunk in parallel}
    \State $part \gets chunks[index]$
    \State $genSum \gets \text{generalSum}(part)$
    \If{$first = True$}
        \State $actions \gets \text{actionItemExtraction}(chunk)$ \Comment{Extract action items}
        \State $combined \gets genSum + actions$
        \State $combinedNumTokens \gets \text{tokenLen}(\text{tokenizer}(genSum + actions))$
        
        \If{$\text{combinedNumTokens} > \text{maxTokens}$} \Comment{Theoretically possible but never true in our testing}
        \State $combined \gets \text{truncateText}(combined)$ 
        \EndIf
        \State $chunkSum \gets \text{generalSum}(combined)$
        \State $chunkSums[index] \gets chunkSum$
    \Else
        \State $chunkSums[index] \gets genSum$
    \EndIf
\EndFor
\State $concatSums \gets \text{concatenate}(chunkSums)$ \Comment{Concatenate summaries after parallel loop completes}
\State $summaryNumTokens \gets \text{tokenLen}(\text{tokenizer}(concatSums))$
\If{$summaryNumTokens > \text{maxTokens}$}
    \State \Return $\text{actionItemDrivenSummary}(concatSums, False, \text{maxTokens})$ \Comment{Recursive call}
\Else
    \State \Return $\text{generalSum}(concatSums)$ \Comment{The action-item-driven summary of $text$}
\EndIf
\Description[Action-Item-Driven Summary algorithm]{Our novel "action-item-driven summary" algorithm as described in Section 3}
\end{algorithmic}
\end{algorithm*}

\section{Results and Analysis}
We first generated meeting summaries without including our action-item extraction technique in order to evaluate our three topic segmentation techniques and recursive algorithm. We evaluate within our own techniques as well as compare to the current state-of-the-art on the AMI dataset using the BART summarizer \cite{shinde_automatic_2022}. Then we compare our summaries with and without action items and show that our action-item-driven summaries contain additional valuable information.

\begin{table*}
\centering
\begin{tabular}{lcccc}
\toprule Topic Segmentation $\downarrow$ \ \ Metric $\rightarrow$
& \textbf{BERTScore} & \textbf{R-1} & \textbf{R-2} & \textbf{R-L} \\
\midrule
\multicolumn{5}{l}{\textbf{General Summaries (Without Action Items)}} \\
\midrule
\textbf{Linear Segmentation (Baseline Technique)} & 63.41 & 38.14 & 8.61 & 19.46 \\
\textbf{Chunked Linear Segmentation} & \textbf{64.77} & \textbf{38.93} & \textbf{9.27} & \textbf{19.63} \\
\textbf{Simple Cosine Segmentation} & 63.91 & 38.49 & 8.61 & 19.46 \\
\textbf{Complex Cosine Segmentation} & 64.48 & 38.92 & 9.24 & 19.47 \\
\midrule
\multicolumn{5}{l}{\textbf{Action-Item-Driven Summaries}} \\ 
\midrule
\textbf{Linear Segmentation (Baseline Technique)} & 63.76 & 35.11 & 8.04 & 18.99 \\
\textbf{Chunked Linear Segmentation} & \textbf{64.98} & \textbf{36.27} & 8.31 & \textbf{19.62} \\
\textbf{Simple Cosine Segmentation} & 64.14 & 35.30 & 8.12 & 19.24 \\
\textbf{Complex Cosine Segmentation} & 64.87 & 36.21 & \textbf{8.32} & 19.61 \\
% your new row without title
\midrule
\textbf{Shinde et al., (2022)} & 60 & 45.2 & 13.3 & - \\
\bottomrule
\end{tabular}
\caption{BERTScore and ROUGE evaluation scores for our machine-generated summaries across 4 different topic segmentation methods on the AMI corpus. This is done separately for both the general summaries (without action items) and the action-item-driven summaries. We also include the scores achieved by the current state-of-the-art model \cite{shinde_automatic_2022}.}
\label{my-label}
\Description[Results Table]{Our results table shows that we outperform the model by \citet{shinde_automatic_2022}, our action-item-driven summaries outperform our general summaries, and our three topic segmentation methods outperform linear segmentation.}
\end{table*}

\subsection{Topic Segmentation Performance}
We evaluate our topic segmentation methods by keeping our recursive algorithm constant and only varying the topic segmentation method. We see in Table 1 that all three of our novel topic segmentation methods outperformed linear segmentation with respect to both the BERTScore and ROUGE metrics. Most notably, with respect to the BERTScore metric, our methods, simple cosine segmentation, complex cosine segmentation, and chunked linear segmentation, outperform linear segmentation by 0.50\% 1.07\% and 1.36\%, respectively for the generated summaries without action items. For the summaries with action items, the improvements over linear segmentation with respect to the BERTScore metric, were 0.38\% 1.11\% and 1.22\%, respectively. 

The complex cosine segmentation technique outperformed the simple cosine segmentation technique by 0.57\% and 0.73\% in terms of the BERTScore metric for the summaries without and with action items, respectively. This was expected because the former was less sensitive to "meaningless" turns as explained in the "Complex Cosine Segmentation" subsection. However, chunked linear segmentation, which does not rely on sentence embeddings and cosine similarity, outperformed all. 

\subsection{Recursive Algorithm Performance}
We also compare the results of our recursive algorithm to the method proposed by \citet{shinde_automatic_2022}. When we both use linear segmentation and the same fine-tuned BART models, but different "recursive" algorithms, our action-item-driven model outperforms the model presented by \citet{shinde_automatic_2022} by approximately 4.98\% in terms of the BERTScore metric. With regard to our general summarization model (without action items), this model still outperformed that presented by \citet{shinde_automatic_2022} by approximately 4.77\%. This means that, regardless of whether or not we include action items, the summaries our model generates are more similar to those of the human reference summaries in terms of their semantic meanings. 

The model by \citet{shinde_automatic_2022} does outperform our model in terms of the ROUGE scores, which measure lexical overlap, but this is expected since we use a truly recursive algorithm that results in the input text and the corresponding sectional summaries being passed into the BART summarizer more times. This would, of course, decrease the lexical overlap between our machine-generated summaries and the human reference summaries. However, it seems that our summaries better match the semantic meaning of the human reference summaries, which was shown to be more important for human judgement by \citet{zhang_bertscore_2020}. 
\begin{table*}
\centering
\begin{tabular}{|p{0.5\textwidth}|p{0.5\textwidth}|}
\hline
\textbf{General Summary (Without Action Items)} & \textbf{Action-Item-Driven Summary} \\
\hline
Marketing Expert, Product Manager, and Industrial Designer are having a conceptual design meeting after lunch. They talk about the most important aspect for remote controls as people want a fancy look and feel. They discuss batteries, the design of the LCD display on the LCD screen, how to distinguish where people have to press the button when they have a flip-top, and how to incorporate voice recognition into the remote control. They agree on keeping the control buttons standardized and checking the financial feasibility. They decide to start with the black and white one and go for a whistle if financially voice recognition is not feasible. The product will have a logo on it just like everything else in a year's time if they get feedback from design fairs. Product manager will go through the end of the end meeting. Marketing Expert shares some information about a remote control that fits into the palm of the hand, made of plastic, with a rubberised cover, and the design is based on the input from the previous meeting.  & 
Marketing Expert, Product Manager, and Industrial Designer are having a conceptual design meeting after lunch. They talk about properties, materials, user-interface and trend-watching. \ul{Marketing Expert says the fashion update which relates to very personal preferences among their subject group.} \ul{There's no rechargeable option for the remote control, so they're going to look into battery options.} Industrial Designer and Marketing Expert are talking about the size of the batteries they need to take into consideration. \ul{Marketing Expert thinks using the standard batteries and the solar charging will detract from the attractiveness of the whole feature.} \ul{Marketing Expert thinks the buttons on the remote should have lights behind the buttons.} Marketing Expert wants to make the basic mold out of plastic but have a rubber cover. \ul{Marketing experts are going to market to guys as much as to women.} Marketing Expert shares with Industrial Designer some information about the design of the LCD display on the LCD screen. Industrial Designer and Marketing Expert are discussing how to incorporate voice recognition into the remote control. \ul{Industrial Designer tells Product Manager they need to get double A or triple A batteries.} Sarah and Marketing Expert are talking about the design of a remote control with a rubberised cover. Industrial Designer tells Marketing Expert they can go for a whistle if voice recognition is not feasible. Product Manager will wrap up the end-of-meeting message.\\
\hline
\end{tabular}
\caption{Comparison between machine-generated General (Without Action Items) and Action-Item-Driven Summaries when both methods employ chunked linear segmentation. The additions in the action-item-driven summary are underlined. AMI Meeting ID: ES2004c}
\label{table:summary}
\Description[Action-Item-Driven Summary vs. General Summary]{A comparison table showing the additional action items included in our machine-generated action-item-driven summary compared to our machine-generated general summary.}
\end{table*}
\subsection{Action-Item-Driven Summary Performance}
As seen in Table 1, our action-item-driven summaries achieve slightly higher BERTScores than our general summaries (without action items), but we consider this difference negligible ($0.21\%$ increase in BERTScore when both use chunked linear segmentation). However, we suspect that the reason for this small difference is that the human reference summaries in the AMI dataset appear to be more action-item-driven that those in the XSUM and SAMSUM datasets which we used to fine-tune our BART model.

The ROUGE scores for our action-item-driven summaries were notably lower than those achieved by our general summaries. For example, when both techniques employ chunked linear segmentation, the ROUGE-1 scores for our general summaries were 1.66\% higher than those for our action-item-driven summaries. This makes sense since, in the action-item-driven summaries, we are deliberately adding words and phrases (action items) that are not included in the human reference summaries; thus, our precision score decreases. However, the slight increase in our BERTScores suggests that we are still capturing the semantic meaning of the reference summaries well.

Table 2 shows example outputs from our general model and our action-item-driven model when both algorithms employ chunked linear segmentation. We underline the additions in the action-item-driven summary and show that our action-item-driven model properly includes relevant action items from the meeting. Consider the following sentence from the action driven summary: "There’s no rechargeable option for the remote
control, so they’re going to look into battery options." This action item is not included in either the general summary or the human reference summary, but it is a relevant and informative action item that adds value to the meeting summary. We also see that this action item is coherent and rich with context. This example and the other sentences underlined in Table 2 serve as evidence that our action-item extraction technique utilizing neighborhood summarization is quite effective. 

\section{Future Research}
In this study, we focused on generating action-item-driven summaries, but there are additional components of a good meeting summary. As noted in our "Introduction" section, decisions made, main topics, tension levels, etc. would also be very informative aspects of a meeting summary. While incorporating these elements into a meeting summary may lower our automated evaluation scores, this does not necessarily mean that the resulting meeting summary would be less useful for human readers. We hope to explore current approaches and develop new algorithms to extract these ideas from a meeting transcript and then incorporate them into a meeting summary. 

While all three of our novel topic segmentation techniques outperformed linear segmentation, our best performance came from chunked linear segmentation, which did not involve calculating any embeddings or cosine similarities. However, the fact that chunked linear segmentation outperformed linear segmentation suggests that we can generate better summaries by minimizing the number of interrupted ideas in the meeting transcript. Thus, we hope to develop a more advanced topic segmentation method that will lead to better generated summaries and outperform chunked linear segmentation. 

Finally, action-item extraction has not been explored in depth and has both a lack of techniques as well as metrics for evaluating these techniques. Thus, we hope to dive deeper into this issue and produce more advanced techniques for accomplishing the two above goals. Nevertheless, our neighborhood summarization algorithm proved very effective in action-item extraction, and we hope to test its performance on other tasks involving context resolution as well (e.g. extracting decisions made from a meeting).

\section{Conclusion}
This study explores a novel method for automatically generating meeting summaries by treating this problem as a fundamentally different one from that of generating dialogue summaries. Action items drive this recursively-generated, abstractive summary of the meeting that achieves approximately 4.98\% higher BERTScores across the AMI corpus than the previous state-of-the-art using the BART summarizer. We introduce novel topic segmentation and action-item extraction algorithms that all improve and add value to the resulting summaries. The recursive approach presented in this paper for generating summaries for different parts and aspects of the meeting transcript can be expanded upon to improve meeting summarization, as well as be generalized and applied to summarizing other genres of text in the future.

\bibliographystyle{ACM-Reference-Format}
\bibliography{draft_first}

\end{document}